\documentclass[letterpaper]{article} 
\usepackage[preprint]{aaai2027}
\usepackage[hyphens]{url}  
\usepackage{graphicx} 
\usepackage{natbib}  
\usepackage{caption} 
\usepackage{algorithm}
\usepackage{algorithmic}
\usepackage{booktabs}
\usepackage{amsmath}
\usepackage{amssymb}

\newcommand{\topk}{\mathrm{TopK}}
\newcommand{\cost}{\mathcal{C}}
\newcommand{\experts}{\mathcal{E}}
\newcommand{\activeset}{\mathcal{U}}

\nocopyright

\title{{DraftExpert}: Expansion-Aware Self-Speculative Decoding for End-Device {MoE} Inference}
\author{Dengke Han}
\affiliations{}

\begin{document}

\maketitle

\begin{abstract}
Large Mixture-of-Experts (MoE) language models are attractive for end-device deployment because only a small subset of experts is active per token, but their routed expert weights often exceed accelerator memory. We target latency-critical single-user settings where routed experts are staged on demand from CPU memory to a GPU or from Flash to a mobile NPU. In this setting, self-speculative decoding faces a new bottleneck: increasing the draft expert set improves accuracy but triggers extra expert loading, while cheap small-footprint drafts have low acceptance; moreover, verifying a multi-token block activates the union of target experts and is no longer close to one target step. We propose \emph{DraftExpert}, an expansion-aware self-speculative decoding framework for expert-offloaded MoE inference. DraftExpert trains one lightweight accelerator-resident draft expert per layer by self-distilling residual, logit/token, and router-agreement signals from the frozen target MoE. At inference time, it uses a fixed-footprint shared+top-1+draft-expert drafter together with confidence--expansion truncation and target-expert prefetching, while final tokens are still exactly verified by the target model. On DeepSeek-V2-Lite and Moonlight-16B-A3B across CPU$\rightarrow$GPU and Flash$\rightarrow$NPU offload, DraftExpert improves decode throughput by $1.45\times$ on average, raises draft acceptance to 84--87\%, and achieves 86--88\% prefetch hit rates.
\end{abstract}


\section{Introduction}

Speculative decoding accelerates autoregressive generation by drafting multiple future tokens and verifying them with the target model in parallel \citep{leviathan2023fast,chen2023accelerating,miao2024specinfer}. Its performance rests on three conditions: \emph{the draft must be cheap}, \emph{parallel verification must be close to one target step}, and \emph{the draft must be accepted often enough}. This paper argues that end-device Mixture-of-Experts (MoE) inference changes the performance paradigm behind all three conditions. When routed experts are offloaded to CPU memory for consumer CPU$\rightarrow$GPU execution or to Flash for mobile Flash$\rightarrow$NPU execution, speculative efficiency is no longer governed mainly by the number of drafted tokens; it is governed by how quickly the draft and verification stages expand the set of experts that must be loaded.

This setting is increasingly important. Large MoE language models are attractive for personal devices because each token activates only a small fraction of experts \citep{shazeer2017outrageously,lepikhin2021gshard,fedus2022switch,du2022glam,jiang2024mixtral,deepseekai2024deepseekv2,liu2025moonlight}. We target latency-critical single-user end-device inference, where interactive generation has limited batching or cross-request cache reuse and therefore exposes per-request expert movement. However, the routed experts dominate the parameter count and often exceed the memory of consumer GPUs or mobile accelerators. A practical runtime therefore keeps routed experts in CPU memory or mobile Flash and loads them on demand, while the rest of the model stays accelerator-resident because it is used every forward pass \citep{xue2024moeinfinity,yi2023edgemoe,kamahori2025fiddler}. Expert movement becomes a first-order latency cost.

Under this expert-offload cost model, existing MoE self-speculative methods struggle with all three conditions. First, the draft is not necessarily cheap. Training-free self-drafters often use the shared path plus router top-$r$ routed experts. Increasing $r$ improves the approximation to the target MoE, but it also makes the drafter load more offloaded experts as the draft grows. Second, verification is not nearly free. A block of $K$ draft tokens may activate the union of many target experts, so the cost of one parallel verify pass grows with expert-set expansion. Third, acceptance is not high when $r$ is kept small. A top-1 or small-$r$ drafter omits routed-expert residuals, causing hidden-state drift, lower token agreement, and weaker router agreement.

The key performance unit therefore shifts from \emph{tokens per target forward} to \emph{accepted tokens per expert-set expansion}. A useful end-device MoE speculative decoder should restore the three classical conditions under this new unit: it should keep draft expansion bounded, make verification expansion predictable and controllable, and recover the accuracy lost by cheap drafting.

We propose \emph{DraftExpert}, an expansion-aware self-speculative decoding framework for expert-offloaded MoE inference. For the draft condition, DraftExpert uses a fixed-footprint shared+top-1+draft-expert path. Each MoE layer receives one lightweight accelerator-resident draft expert, while the original target MoE remains frozen. The draft expert is trained to compensate for the routed experts omitted by top-1 drafting, so the drafter can improve accuracy without increasing $r$ or loading more experts.

For the acceptance condition, DraftExpert uses self-distillation from the full target MoE. Residual, logit, and token losses teach the draft expert to recover missing expert behavior and improve token agreement. For the verification condition, DraftExpert adds router-agreement distillation, making cheap draft hidden states more predictive of the target experts that verification will use. These predictions enable two exact runtime policies: a confidence--expansion controller truncates drafts when a low-confidence candidate would introduce too many new verifier experts, and draft-router prefetching moves likely verifier experts before the target reaches them. All emitted draft tokens remain verified by the frozen full MoE.

This paper makes three contributions:
\begin{itemize}
    \item We formulate expert-offloaded MoE speculative decoding as an \emph{expert-set expansion} problem and show how expansion breaks the three classical conditions for speculative speedup.
    \item We introduce DraftExpert, a fixed-footprint shared+top-1+draft-expert self-drafter that reduces draft-stage expert loading while self-distillation restores acceptance.
    \item We use router-agreement distillation to predict verifier expert sets, enabling expansion-aware dynamic truncation and draft-router expert prefetching for exact target verification.
\end{itemize}

\section{Background and Motivation}

\subsection{Speculation Conditions Under Expert Offload}

\begin{figure*}[t]
\centering
\includegraphics[width=\textwidth]{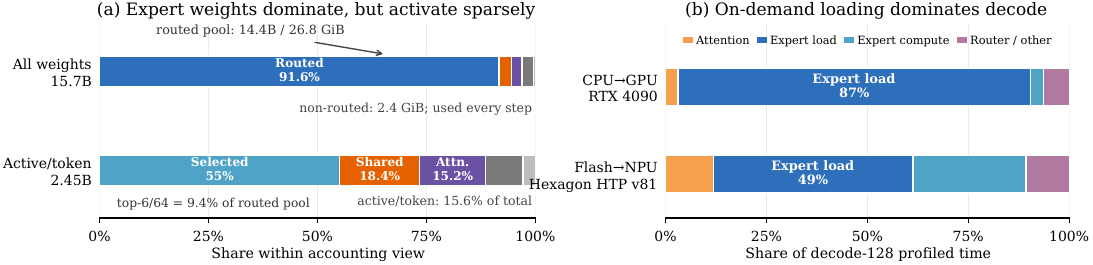}
\caption{Motivation for expert-offloaded MoE inference on end devices. (a) Routed experts dominate DeepSeek-V2-Lite weight storage, but each token activates only a small selected subset; non-routed weights are much smaller and used every forward pass. (b) With non-routed weights resident and routed experts loaded on demand, expert loading dominates decode-128 profiled time in both CPU$\rightarrow$GPU and Flash$\rightarrow$NPU settings. Flash$\rightarrow$NPU uses llama.cpp on a Hexagon HTP v81 NPU and excludes runtime API gap.}
\label{fig:offload-motivation}
\end{figure*}

Figure~\ref{fig:offload-motivation}(a) shows the parameter asymmetry that makes expert offloading both necessary and natural on memory-constrained end devices. In DeepSeek-V2-Lite, routed experts account for 14.4B parameters, or 91.6\% of the model weights. In BF16, this routed pool alone occupies 26.8~GiB, while all non-routed weights together occupy only 2.4~GiB. At the same time, the routed pool is sparsely activated: each token selects only top-6 out of 64 routed experts per MoE layer, i.e., 9.4\% of the routed pool. By contrast, attention, shared experts, dense layers, routers, norms, and output projection are exercised on every forward pass.

Therefore, the most feasible way to run large MoE models on end devices is to keep non-routed weights resident on the accelerator and place the large routed-expert pool in slower memory: CPU DRAM for consumer GPUs or Flash storage for mobile NPU/GPU execution. Selected experts are loaded on demand. This deployment enables models whose total parameters exceed accelerator memory, but it also changes the performance model: Figure~\ref{fig:offload-motivation}(b) shows that expert movement, rather than arithmetic, becomes the dominant decode bottleneck.

Speculative decoding is fast only when three conditions hold: drafting is cheap, verifying a block is close to one target step, and the accepted prefix is long enough to amortize the draft work. In dense models, these conditions are often analyzed in units of target forward passes. In expert-offloaded MoE inference, the natural unit is different. Because routed experts live outside accelerator memory and are loaded on demand, the cost of both drafting and verification depends on the unique experts touched by a token block.

For an MoE layer $l$, let the target router select
\begin{equation}
    \topk_l(h) \subseteq \{1,\ldots,N_l\}, \quad |\topk_l(h)|=k_l.
\end{equation}
For a block $B$, define its expert set and offload cost as
\begin{equation}
    \activeset_l(B)=\bigcup_{v\in B}\topk_l(h_{l,v}), \quad
    \cost(B)=\sum_l\sum_{e\in\activeset_l(B)} c_{l,e},
\end{equation}
where $c_{l,e}$ captures expert loading and execution. The central bottleneck is \emph{expert-set expansion}: each additional token may introduce newly loaded experts. Because Figure~\ref{fig:offload-motivation}(b) shows that expert loading dominates decode, expansion directly turns into latency.

\subsection{Condition 1: Drafting Must Stay Cheap}

MoE self-drafting avoids a separate draft model by reusing the target model itself. A common strategy evaluates the shared path and router-selected top-$r$ routed experts. This creates a direct acceptance--cost trade-off. Larger $r$ better approximates the full target MoE and usually improves draft accuracy, but it also makes the drafter load more offloaded experts. Thus the drafter becomes less cheap exactly when it becomes more accurate.

\begin{figure}[t]
\centering
\includegraphics[width=\columnwidth]{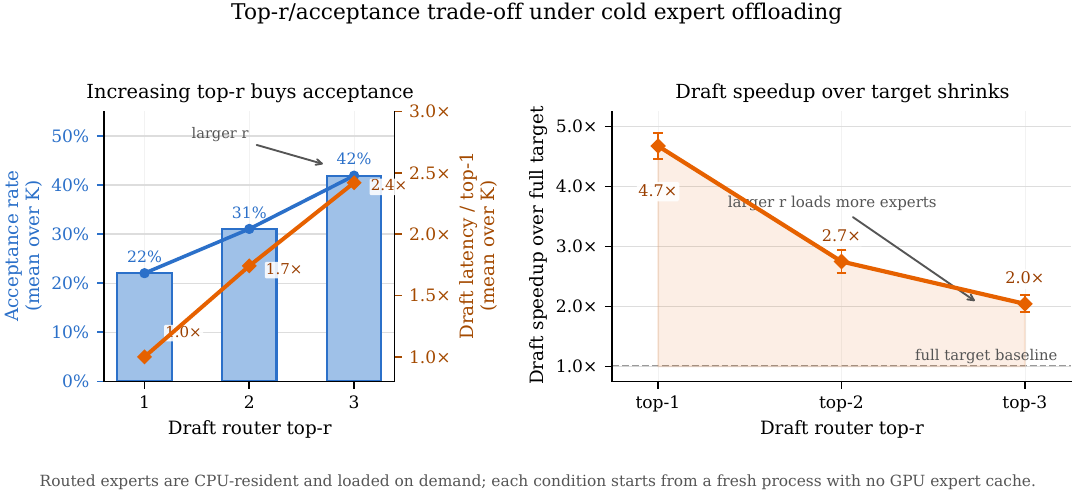}
\caption{Top-$r$ acceptance--cheapness trade-off under cold expert offloading. Increasing router top-$r$ improves draft acceptance, but it also makes drafting less cheap: draft latency rises relative to top-1, while draft speedup over the full target falls from 4.7$\times$ to 2.0$\times$.}
\label{fig:topr-acceptance-cost}
\end{figure}

\begin{figure}[t]
\centering
\includegraphics[width=\columnwidth]{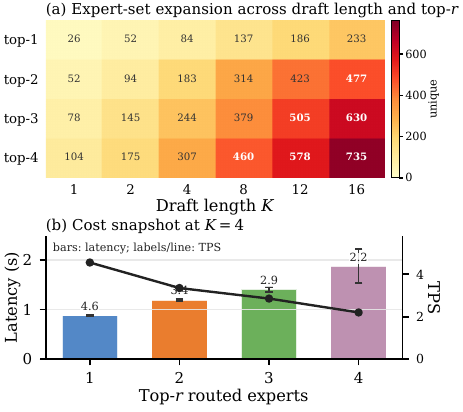}
\caption{Draft-side expert-set expansion. Shared+top-$r$ self-drafters improve approximation by loading more routed experts, so draft cost grows with $r$ and draft length.}
\label{fig:draft-topr-tradeoff}
\end{figure}

Figure~\ref{fig:topr-acceptance-cost} quantifies this trade-off in our cold expert-offload setting. Averaged over draft lengths, increasing $r$ raises acceptance from 22\% to 31\% and 42\%, but it also increases draft latency relative to top-1. More importantly, the drafter's speedup over the full target drops from 4.7$\times$ for top-1 to 2.7$\times$ for top-2 and 2.0$\times$ for top-3. Thus naive top-$r$ self-drafting buys acceptance by spending the cheap-draft budget on additional expert loading. Figure~\ref{fig:draft-topr-tradeoff} further illustrates the underlying mechanism: as draft length or $r$ grows, the cumulative unique routed experts used by drafting expands rapidly. Therefore, recovering the cheap-draft condition requires a fixed-footprint drafter: accuracy should come from a small accelerator-resident learned module, not from loading more routed experts.

\subsection{Condition 2: Parallel Verification Must Be Controlled}

\begin{figure}[t]
\centering
\includegraphics[width=\columnwidth]{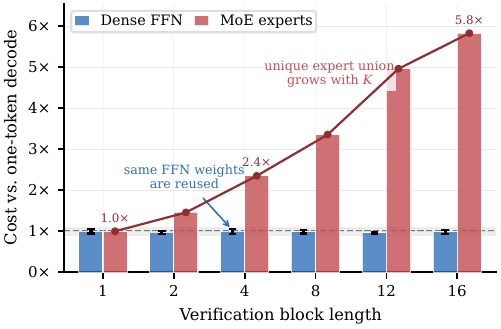}
\caption{Dense versus MoE verification under offloaded weights. Dense FFN verification reuses the same weights across tokens, while MoE verification becomes more expensive as longer blocks activate more unique routed experts.}
\label{fig:dense-vs-moe-verify}
\end{figure}

Dense speculative decoding benefits from the fact that verifying multiple tokens reuses the same dense FFN weights. Offloaded MoE verification is different: a verification block may route different tokens to different experts, so one parallel pass can still load many unique experts. Figure~\ref{fig:dense-vs-moe-verify} compares dense FFN offload with MoE expert offload after normalizing each runtime by its own one-token decode cost. Dense verification remains close to one step, while MoE verification grows with block length.

\begin{figure}[t]
\centering
\includegraphics[width=\columnwidth]{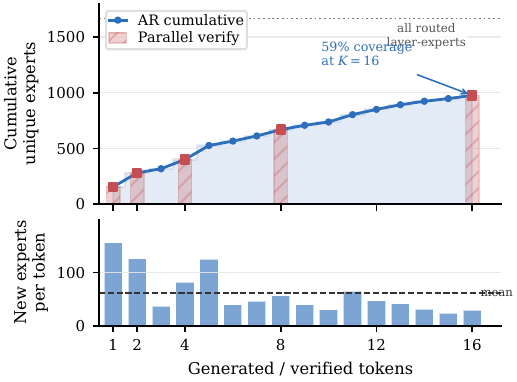}
\caption{Verify-side expert-set expansion. Multi-token decode or verification activates a growing union of routed layer-experts, so parallel verification is not free when target experts are loaded from CPU memory or Flash.}
\label{fig:verify-expansion}
\end{figure}

For a verification block $D=(v_1,\ldots,v_K)$, the verifier may access
\begin{equation}
    \activeset_l(D)=\bigcup_{i=1}^{K}\topk_l(h_{l,v_i}^{\star}).
\end{equation}
Figure~\ref{fig:verify-expansion} directly measures this expansion. The verify condition should therefore be reframed as controlling marginal expert expansion, not merely maximizing parallel block length.

\subsection{Condition 3: Acceptance Requires Learned Drafting}

One might keep draft and verify expansion small by using only shared+top-1 drafting and short fixed blocks. However, this sacrifices the third condition: high acceptance. Top-1 drafting omits the residual contribution of other target experts, causing hidden-state drift and lower next-token agreement. Simply spending more expert-loading cost on naive shared+top-$r$ drafting does not solve the problem either: Figure~\ref{fig:topr-acceptance-cost} shows that even the much more expensive top-3 path remains below 50\% acceptance. In other words, naive self-drafting pays a large movement cost while still rejecting most candidates, which is close to guessing from the verifier's perspective.

Low acceptance also weakens router agreement, which matters for verification because offload runtimes often use the drafter's router to predict or prefetch verifier experts. If the cheap drafter's hidden states drift, expert prediction becomes inaccurate, prefetch hit rate falls, and verification stalls remain. This motivates a dedicated self-distilled draft expert: acceptance should be recovered by learning the missing routed-expert residual, not by loading a larger top-$r$ expert set during drafting.

These observations suggest an expansion-aware design principle: optimize accepted tokens per newly loaded expert. A method should (i) bound draft expert-set expansion, (ii) recover acceptance without increasing $r$, and (iii) make verifier expert expansion predictable enough for truncation and prefetching.

\section{DraftExpert Method}

\subsection{Overview}

DraftExpert restores the three speculative decoding conditions under an expert-set expansion cost model. To make the draft cheap, it replaces shared+top-$r$ self-drafting with a fixed-footprint shared+top-1+draft-expert path. To keep acceptance high, it self-distills the draft experts from the frozen full MoE so the cheap path recovers information from omitted routed experts. To make verification close to cheap again, it improves router agreement and uses the predicted verifier expert set for cost-aware truncation and prefetching.

The deployment assumption is that target routed experts are offloaded to CPU memory or Flash and loaded on demand, while the non-expert components and DraftExpert's added draft experts are accelerator-resident. The original target MoE is frozen. In exact mode, every emitted draft token is verified by the full target MoE; truncation and prefetching change only how many candidates are proposed and when experts are moved.

\begin{figure*}[t]
\centering
\includegraphics[width=0.98\textwidth]{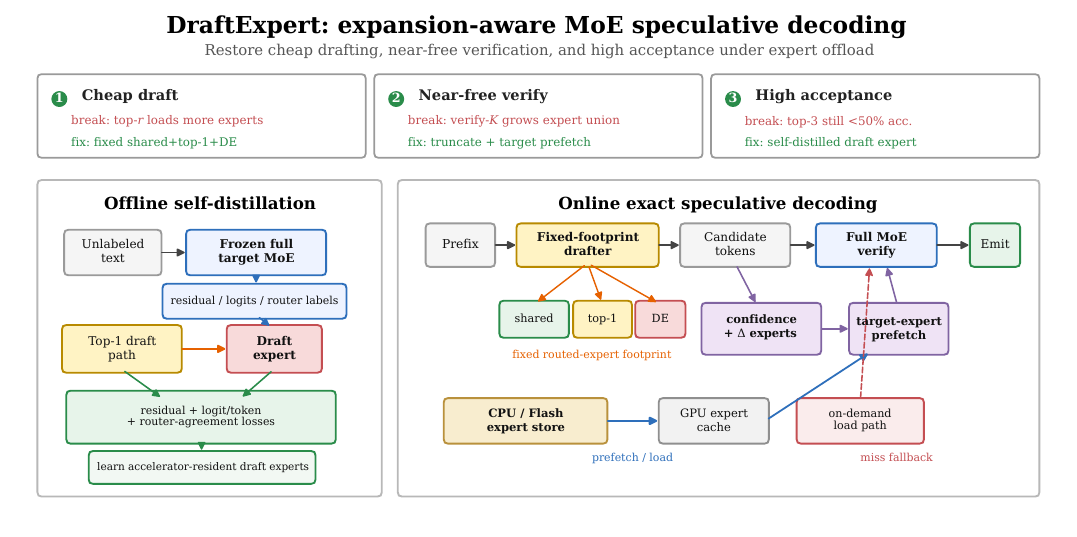}
\caption{Expansion-aware DraftExpert overview. Expert-offloaded MoE breaks the three classical speculative decoding conditions through expert-set expansion: top-$r$ drafting is expensive yet still yields low acceptance, parallel verification activates a growing target-expert union, and naive small-footprint drafting drifts from the verifier. DraftExpert restores these conditions with a fixed-footprint shared+top-1+draft-expert path, self-distillation with router agreement, and expansion-aware truncation plus target-expert prefetching for exact verification.}
\label{fig:draftexpert-architecture}
\end{figure*}

\subsection{Fixed-Footprint Drafting}

Let $h_l$ be the hidden state entering MoE layer $l$, $S_l$ be the shared expert path, and $\mathcal{R}_l(h_l)$ be the original router distribution. The full target MoE output is
\begin{equation}
    y_l^{\star}=S_l(h_l)+\sum_{e\in \topk_l(h_l)} g_{l,e}E_{l,e}(h_l),
\end{equation}
where $E_{l,e}$ are frozen routed experts. A shared+top-$r$ drafter reduces cost by using fewer routed experts than the target, but its cost still grows with $r$ and with the unique experts touched by the draft block.

DraftExpert instead adds one draft expert $E_l^{\mathrm{d}}$ per MoE layer and drafts with
\begin{equation}
\begin{aligned}
    y_l^{\mathrm{d}} &= S_l(h_l) + g_{l,e_1}E_{l,e_1}(h_l) + \gamma_l E_l^{\mathrm{d}}(h_l), \\
    e_1 &= \arg\max_e \mathcal{R}_l(h_l)_e .
\end{aligned}
\end{equation}
Only the draft experts and optional scales are trainable. This keeps routed-expert usage close to top-1 while adding a small accelerator-resident module that can learn the missing top-$k$ information.

\subsection{Self-Distillation for Accuracy and Router Agreement}

Self-distillation has two roles. The first is to improve draft accuracy without increasing $r$. For a teacher-forced sequence, define the missing routed-expert residual after removing the shared and top-1 paths:
\begin{equation}
    m_{l,t}^{\star}=y_{l,t}^{\star}-S_l(h_{l,t}^{\star})-g_{l,e_1}E_{l,e_1}(h_{l,t}^{\star}).
\end{equation}
The draft expert is trained to approximate this residual,
\begin{equation}
    \mathcal{L}_{\mathrm{res}}=\sum_{l,t}\left\|\gamma_lE_l^{\mathrm{d}}(h_{l,t}^{\mathrm{d}})-m_{l,t}^{\star}\right\|_2^2.
\end{equation}
We also distill final logits and top-1 target tokens:
\begin{equation}
    \mathcal{L}_{\mathrm{logit}}=T^2\sum_t\mathrm{KL}\left(\mathrm{softmax}(z_t^{\star}/T)\Vert\mathrm{softmax}(z_t^{\mathrm{d}}/T)\right),
\end{equation}
\begin{equation}
    \mathcal{L}_{\mathrm{tok}}=\sum_t \mathrm{CE}\left(\arg\max z_t^{\star},\mathrm{softmax}(z_t^{\mathrm{d}})\right).
\end{equation}

The second role is verify-side expert prediction. If the draft hidden states drift from the target path, the draft router becomes a poor predictor of verifier experts. We therefore include router-agreement supervision,
\begin{equation}
    \mathcal{L}_{\mathrm{router}}=\sum_{l,t}\mathrm{KL}\left(\mathcal{R}_l(h_{l,t}^{\star})\Vert \mathcal{R}_l(h_{l,t}^{\mathrm{d}})\right),
\end{equation}
and optionally track top-$k$ expert-set agreement as a validation metric. The complete objective is
\begin{equation}
    \mathcal{L}=\mathcal{L}_{\mathrm{logit}}+\alpha\mathcal{L}_{\mathrm{tok}}+\beta\mathcal{L}_{\mathrm{res}}+\eta\mathcal{L}_{\mathrm{router}}.
\end{equation}
All supervision comes from unlabeled text and the frozen target MoE.

\subsection{Cost-Aware Dynamic Truncation}

For the $i$-th draft token, let $q_i$ be a confidence score from the draft logits. Let $\hat{\experts}_{l,i}$ be the verifier experts predicted by the draft router for layer $l$. DraftExpert maintains a predicted block-level verifier set
\begin{equation}
    \hat{\activeset}_i=\bigcup_{j\le i}\bigcup_l \hat{\experts}_{l,j},
\end{equation}
with marginal expansion
\begin{equation}
    \Delta_i=|\hat{\activeset}_i|-|\hat{\activeset}_{i-1}|.
\end{equation}
The controller stops before appending token $i$ when the token is unlikely to be accepted and likely to add verifier cost:
\begin{equation}
    q_i < \tau \quad \text{and} \quad \Delta_i > B_i,
\end{equation}
where $B_i$ can be a fixed layer-expert budget or a budget derived from the current expert cache. This policy is exact because discarded candidates are never emitted, and all emitted candidates are verified by the target MoE.

\subsection{Draft-Router Expert Prefetch}

The same router-aligned draft states support prefetching. As soon as draft token $i$ is kept, DraftExpert launches non-blocking prefetches for newly predicted verifier experts $\hat{\experts}_{l,i}$ that are not already cached or in flight. These transfers overlap with the remaining draft-token computation; verification only waits for unfinished predicted transfers when it reaches the target pass, and any missed expert is still loaded on demand. Prefetching does not change model semantics: if a prefetched expert is used, verification avoids an on-demand stall; if it is unused, the cost is wasted bandwidth; if a required expert is missed, the verifier loads it on demand. Router-agreement distillation is therefore important because it directly affects prefetch hit rate and wasted movement.

\begin{algorithm}[t]
\caption{Exact DraftExpert Decoding}
\label{alg:draftexpert}
\begin{algorithmic}[1]
\STATE $D\leftarrow\emptyset$, $\hat{\activeset}\leftarrow\emptyset$, $\mathcal{P}\leftarrow\emptyset$
\STATE $\mathcal{C}\leftarrow$ current accelerator expert cache
\FOR{$i=1$ to $K_{\max}$}
    \STATE Draft $v_i$ with shared+top-1+draft-expert path
    \STATE Compute confidence $q_i$ and predicted verifier experts $\hat{\experts}_{l,i}$
    \STATE $\hat{\activeset}_i\leftarrow \hat{\activeset}\cup\bigcup_l\hat{\experts}_{l,i}$
    \STATE $\Delta_i\leftarrow |\hat{\activeset}_i|-|\hat{\activeset}|$
    \IF{$q_i<\tau$ and $\Delta_i>B_i$}
        \STATE Stop drafting before appending $v_i$
        \STATE \textbf{break}
    \ENDIF
    \STATE Append $v_i$ to $D$ and set $\hat{\activeset}\leftarrow\hat{\activeset}_i$
    \STATE Launch async prefetch for $\hat{\activeset}_i\setminus(\mathcal{C}\cup\mathcal{P})$
    \STATE Add launched experts to in-flight set $\mathcal{P}$
\ENDFOR
\STATE Wait for useful in-flight prefetches; load misses on demand
\STATE Verify all tokens in $D$ with the frozen full target MoE
\STATE Accept the target-consistent prefix and continue decoding
\end{algorithmic}
\end{algorithm}

\section{Experimental Evaluation}

We evaluate DraftExpert on two expert-offloaded MoE targets, DeepSeek-V2-Lite (DS) \citep{deepseekai2024deepseekv2} and Moonlight-16B-A3B (ML) \citep{liu2025moonlight}. The experiments are organized around the three conditions in Section~2: a cheap drafter, controlled verification, and high acceptance. Unless noted otherwise, all throughput numbers are \emph{decode-stage} tokens per second after prefill, and speedups are normalized to autoregressive (AR) expert-offloaded decoding on the same model and platform.

\subsection{Setup and Metrics}

We use the same models, prompts, and decoding policies across two end-device memory hierarchies. In the consumer-GPU setting, routed experts are resident in host CPU memory and copied to a discrete GPU on demand. In the mobile setting, CPU, GPU, and NPU share physical DRAM, but DRAM is too small to keep all routed experts resident; routed experts are stored in Flash and staged into shared memory for NPU execution. Mobile measurements use llama.cpp \citep{ggerganov2023llamacpp} with Q4\_0 quantized model weights on the Hexagon HTP NPU. Attention, embeddings, output heads, shared experts, routers, and DraftExpert's added draft experts remain accelerator-resident. Routed target experts are not kept in a persistent accelerator cache unless an experiment explicitly studies the resident-memory reference.

\begin{table}[t]
\centering
\scriptsize
\setlength{\tabcolsep}{2.2pt}
\begin{tabular}{@{}p{0.21\columnwidth}p{0.71\columnwidth}@{}}
\toprule
Item & Configuration \\
\midrule
CG & NVIDIA GeForce RTX 4090, 24 GB memory; BF16 PyTorch 2.11.0+cu128; CPU-resident routed experts loaded on demand to GPU. \\
MN device & Xiaomi 2509FPN0BC / MIX Flip 2 (popsicle/canoe), Android 16 API 36, MIUI V816, Snapdragon 8 Elite (SM8850, TSMC N3E). \\
MN NPU & Hexagon HTP v81 fused AI accelerator; INT8/INT4 peak about 73/146 TOPS; INT4, INT8, INT16, FP16, BF16 support; micro-tile inference; about 8 MB on-chip SRAM. \\
MN runtime & llama.cpp with Q4\_0 model weights. Routed experts are Flash-resident and staged into shared CPU/GPU/NPU memory on demand. The Flash-staged AR decode baseline averages 10.18 TPS; if experts are already memory-resident, the DS Q4\_0 AR reference reaches about 14.36 TPS. \\
Prompts & Chat, GSM8K, MBPP, summarization, and instruction following. \\
Models & DS and ML each use 27 layers, 64 routed + 2 shared experts, and top-6 active routed experts. \\
\bottomrule
\end{tabular}
\caption{Evaluation platforms. CG denotes CPU$\rightarrow$GPU expert offload, and MN denotes Flash$\rightarrow$NPU expert offload. Mobile throughput is reported for the Q4\_0 llama.cpp NPU runtime; the resident-memory value is an AR reference rather than the offloaded baseline.}
\label{tab:platform}
\end{table}

We compare three method groups. \textbf{AR offload} decodes with the frozen target MoE and loads routed experts on demand. \textbf{Shared+top-$r$} is the strongest training-free self-drafting baseline after sweeping $r\in\{1,2,3\}$. \textbf{DraftExpert} uses the fixed-footprint shared+top-1+draft-expert path, router-agreement distillation, confidence--expansion truncation, and target-expert prefetch. All final tokens are verified by the frozen target MoE, so DraftExpert preserves exact target outputs.

\subsection{Decode Throughput}

Table~\ref{tab:main-results} reports decode-stage throughput. DraftExpert improves average decode TPS by $1.45\times$ over AR offload across the two models and two platforms, but the gains are intentionally not uniform across platforms. CG uses BF16 weights over PCIe into an RTX 4090, whereas MN uses Q4\_0 weights staged from Flash into a shared-memory Hexagon HTP runtime; both absolute TPS and normalized speedup therefore reflect different loading bandwidths, quantization formats, and accelerator compute ratios. On MN, the DS Q4\_0 Flash-staged AR baseline reaches 10.18 TPS, while the resident-memory AR reference reaches about 14.36 TPS; this gap quantifies the loading headroom before speculative gains are applied.

\begin{table}[t]
\centering
\scriptsize
\setlength{\tabcolsep}{2.4pt}
\begin{tabular}{@{}llrrrr@{}}
\toprule
Model & Plat. & AR TPS & top-$r$ TPS & DE TPS & Speedup \\
\midrule
DS & CG & 2.19 & 1.19 & 2.99 & 1.36$\times$ \\
DS & MN & 10.18 & 4.80 & 15.47 & 1.52$\times$ \\
ML & CG & 1.94 & 1.01 & 2.50 & 1.29$\times$ \\
ML & MN & 8.50 & 3.82 & 13.69 & 1.61$\times$ \\
\midrule
Avg. & -- & -- & -- & -- & 1.45$\times$ \\
\bottomrule
\end{tabular}
\caption{Decode-stage exact speculative throughput. AR TPS is the expert-offloaded autoregressive baseline. top-$r$ is the best training-free shared+top-$r$ exact self-drafting setting, including drafting, verification, and repair. DE is DraftExpert with truncation and prefetch.}
\label{tab:main-results}
\end{table}

The main trend is consistent, but its magnitude differs across CG and MN. Training-free shared+top-$r$ improves draft accuracy but is slower than AR once exact verification and repair are included, because it pays extra expert movement before many candidates are rejected. DraftExpert keeps the draft footprint close to top-1 and recovers acceptance through self-distillation, so it converts expert-loading savings into end-to-end decode speedup. The larger MN gains reflect the Flash-staging bottleneck despite Q4\_0 quantization, while CG gains are more compute- and PCIe-balanced.

\subsection{Draft-Side Cost and Acceptance}

The draft-side experiment isolates the first condition: drafting must remain cheap. For draft length $K=4$, Table~\ref{tab:draft-side} reports the cumulative unique routed layer-experts touched by the drafter, accepted/drafted ratio, and draft cost normalized to shared+top-1 on the same platform. Increasing $r$ raises acceptance, but the unique expert set expands much faster. DraftExpert keeps the expert footprint close to top-1 while reaching high acceptance.

\begin{table}[t]
\centering
\scriptsize
\setlength{\tabcolsep}{2.2pt}
\begin{tabular}{@{}llrrrr@{}}
\toprule
Model & Drafter & Unique & Acc. & Cost-CG & Cost-MN \\
\midrule
DS & top-1 & 84 & 22.5\% & 1.00 & 1.00 \\
DS & top-2 & 183 & 31.4\% & 1.36 & 1.47 \\
DS & top-3 & 244 & 45.9\% & 1.59 & 1.83 \\
DS & DraftExpert & 88 & 85.0\% & 1.05 & 1.08 \\
\midrule
ML & top-1 & 86 & 21.8\% & 1.00 & 1.00 \\
ML & top-2 & 186 & 30.2\% & 1.34 & 1.45 \\
ML & top-3 & 248 & 44.0\% & 1.61 & 1.86 \\
ML & DraftExpert & 90 & 83.0\% & 1.06 & 1.09 \\
\bottomrule
\end{tabular}
\caption{Draft-side acceptance--movement trade-off at $K=4$. Cost is draft latency normalized to shared+top-1 on the same model and platform. MN grows faster because additional experts are staged from Flash.}
\label{tab:draft-side}
\end{table}

\subsection{Verify-Side Control}

The verify-side experiment isolates the second condition: verifying a block should be close to one target step. In expert-offloaded MoE, fixed long verification blocks activate a growing union of target experts. Table~\ref{tab:verify-side} compares fixed-$K$ verification with confidence--expansion truncation and prefetch. The reported cost is normalized to one-token AR decode on the same model and platform.

\begin{table}[t]
\centering
\scriptsize
\setlength{\tabcolsep}{2.4pt}
\begin{tabular}{@{}llrrr@{}}
\toprule
Model & Policy & Cost-CG & Cost-MN & Hit/Waste \\
\midrule
DS & Fixed $K$ & 2.35$\times$ & 3.10$\times$ & -- \\
DS & Conf.+exp. & 1.48$\times$ & 1.70$\times$ & -- \\
DS & + Prefetch & 1.25$\times$ & 1.35$\times$ & 88/12\% \\
\midrule
ML & Fixed $K$ & 2.55$\times$ & 3.30$\times$ & -- \\
ML & Conf.+exp. & 1.52$\times$ & 1.76$\times$ & -- \\
ML & + Prefetch & 1.28$\times$ & 1.40$\times$ & 86/14\% \\
\bottomrule
\end{tabular}
\caption{Verify-side cost control. Confidence--expansion truncation avoids low-value candidates that would add many verifier experts, and prefetch hides part of the remaining expert-load stall.}
\label{tab:verify-side}
\end{table}

These results explain why the final throughput gain is moderate rather than equal to the full resident-memory upper bound. DraftExpert reduces and overlaps expert movement, but exact verification still exposes misses, prefetch waste, and non-expert runtime overhead.

\subsection{Training Objective Ablation}

Finally, Table~\ref{tab:training-ablation} keeps the training ablation next to the loss description. The metrics are platform-independent because the same frozen target and DraftExpert checkpoint are evaluated on both CG and MN; their platform-specific latency impact appears in Tables~\ref{tab:main-results}--\ref{tab:verify-side}. Residual distillation teaches the draft expert to approximate omitted routed-expert residuals. Logit and token losses improve acceptance. Router-agreement distillation improves prediction of target experts during verification, which is required for high prefetch hit rate.

\begin{table}[t]
\centering
\scriptsize
\setlength{\tabcolsep}{2.6pt}
\begin{tabular}{@{}llrrrr@{}}
\toprule
Model & Loss & KL & Acc. & Router & Hit \\
\midrule
DS & Top-1 & 1.63 & 47.7\% & 0.62 & 61\% \\
DS & + Residual & 0.15 & 79.5\% & 0.68 & 69\% \\
DS & + Logit/token & 0.07 & 87.1\% & 0.72 & 73\% \\
DS & + Router & 0.07 & 87.0\% & 0.86 & 88\% \\
\midrule
ML & Top-1 & 1.70 & 45.0\% & 0.61 & 59\% \\
ML & + Residual & 0.18 & 76.5\% & 0.66 & 67\% \\
ML & + Logit/token & 0.08 & 84.0\% & 0.70 & 71\% \\
ML & + Router & 0.08 & 84.5\% & 0.85 & 86\% \\
\bottomrule
\end{tabular}
\caption{Training objective ablation. Router denotes the router-agreement loss. Residual/logit/token terms recover draft acceptance, while router agreement improves prefetch hit rate.}
\label{tab:training-ablation}
\end{table}

Overall, DraftExpert turns expert-set expansion into the optimization target: it keeps draft movement nearly fixed, truncates verify blocks when marginal expert expansion is too high, and prefetches the experts most likely to be needed by exact verification.

\section{Related Work}

\paragraph{Speculative decoding.}
Speculative decoding accelerates language-model generation by drafting multiple candidate tokens and verifying them in one target-model pass \citep{leviathan2023fast,chen2023accelerating,miao2024specinfer}. Subsequent systems improve the drafter with token trees, extra decoding heads, feature-level prediction, or early-exit self-drafting \citep{cai2024medusa,li2024eagle,zhang2024draftverify,elhoushi2024layerskip}. Most analyses are shaped by dense-model deployments where verification can reuse the same weights across a token block. DraftExpert focuses on the offloaded MoE case, where the draft stage and verify stage are both governed by unique expert movement.

\paragraph{MoE models and MoE speculation.}
Sparse MoE models increase parameter count while activating only a subset of experts per token \citep{shazeer2017outrageously,lepikhin2021gshard,fedus2022switch,du2022glam,jiang2024mixtral,dai2024deepseekmoe,deepseekai2024deepseekv2,liu2025moonlight}. MoE-specific speculative decoding methods exploit routing structure, expert budgets, or target-model components to reduce draft cost \citep{mcdanel2026moespec,huang2025moesd,zheng2026ssmoe,bang2026specmoe}. DraftExpert addresses the offloaded version of this problem: using more routed experts in the drafter can improve acceptance but directly increases CPU/Flash-to-accelerator transfers.

\paragraph{Expert offloading and prefetching.}
Expert offloading systems move routed experts through heterogeneous memory when accelerator memory is insufficient. Prior systems study expert caching, storage-hierarchy partitioning, CPU--GPU orchestration, quantization, scheduling, and prefetching \citep{xue2024moeinfinity,yi2023edgemoe,kamahori2025fiddler,wang2025specmoeoff,chen2025spmoe,wang2025moespeq,li2026moespac}. DraftExpert is complementary: it trains the drafter to have better router agreement, then uses those predictions for prefetching. The key distinction is that the learned drafter is designed to be both cheap and predictive.

\paragraph{Dynamic speculation control.}
Dynamic speculation methods adapt draft length based on confidence, utility, or serving conditions \citep{saxena2025utility,cheng2026dspark}. DraftExpert specializes this principle for offloaded MoE by combining token confidence with predicted marginal expert expansion. The controller stops drafts when an additional token is both uncertain and likely to add verification expert-loading cost.

\paragraph{Distillation for drafters.}
Many speculative systems train a separate draft model or align a smaller model to the target, often building on knowledge distillation or self-distilled draft heads \citep{hinton2015distilling,cai2024medusa,li2024eagle}. DraftExpert instead distills the frozen target MoE into small per-layer draft experts while reusing the original model. The loss is designed for both sides of the system: residual/logit/token terms improve draft acceptance, while router-agreement terms improve verify-time expert prediction and prefetching.

\begin{table}[t]
\centering
\footnotesize
\setlength{\tabcolsep}{2pt}
\begin{tabular}{@{}p{0.27\columnwidth}p{0.31\columnwidth}p{0.34\columnwidth}@{}}
\toprule
Method family & Main idea & Difference from DraftExpert \\
\midrule
Standard SD & External cheap drafter & Extra model stresses memory \\
MoE-SD / MoE-Spec & MoE-aware speculation or budgets & Does not learn fixed-footprint experts \\
SS-MoE / SpecMoE & Self-drafting from target experts & Top-$r$ draft footprint grows \\
Offload systems & Prefetch/schedule target experts & Need accurate cheap router lookahead \\
Dynamic control & Adapt speculation length & Usually not expert-expansion aware \\
DraftExpert & Distilled draft experts + cost gate & Optimizes both draft and verify costs \\
\bottomrule
\end{tabular}
\caption{Positioning of DraftExpert relative to speculative decoding and MoE offloading work.}
\label{tab:related}
\end{table}

\section{Conclusion}

This paper reframes speculative decoding for end-device MoE inference as a two-sided offload problem. On the draft side, shared+top-$r$ self-drafting improves acceptance by loading more routed experts, making the drafter less cheap. On the verify side, parallel verification activates a growing union of target experts, and prefetching those experts requires router predictions that remain accurate despite cheap drafting.

DraftExpert addresses both sides with a fixed-footprint shared+top-1+draft-expert drafter trained by self-distillation. Residual, logit, and token losses recover draft accuracy without increasing routed-expert usage; router-agreement loss improves prediction of verifier experts. The resulting predictions drive cost-aware dynamic truncation and draft-router prefetching while exact target verification is preserved. This aligns speculative decoding with the memory hierarchy of consumer CPU$\rightarrow$GPU and mobile Flash$\rightarrow$NPU MoE deployment.

\bibliography{refs}


\end{document}